\documentclass{article} 
\usepackage{mypaper}
\usepackage{iclr2022_conference,times}

\title{Robust and Scalable SDE Learning:\\ A Functional Perspective}

\author{%
  Scott Cameron \\
  Oxford University, 
  InstaDeep Ltd.\\
  United Kingdom\\
  \texttt{s.cameron@instadeep.com} \\
  \And
  Tyron Cameron\\
  Discovery Insure\\
  South Africa\\
  \And
  Arnu Pretorius\\
  InstaDeep Ltd.\\
  South Africa\\
  \And
  Stephen Roberts\\
  Oxford University\\
  United Kingdom\\
}

\iclrfinalcopy 
\begin{document}

\maketitle

\begin{abstract}
Stochastic differential equations provide a rich class of flexible generative
models, capable of describing a wide range of spatio-temporal processes. A host
of recent work looks to learn data-representing SDEs, using neural networks and
other flexible function approximators. Despite these advances, learning remains
computationally expensive due to the sequential nature of SDE integrators. In
this work, we propose an importance-sampling estimator for probabilities of
observations of SDEs for the purposes of learning. Crucially, the approach we
suggest does not rely on such integrators. The proposed method produces
lower-variance gradient estimates compared to algorithms based on SDE
integrators and has the added advantage of being embarrassingly parallelizable.
This facilitates the effective use of large-scale parallel hardware for massive
decreases in computation time.
\end{abstract}

\section{Introduction}

Stochastic differential equations~(SDEs) are a natural extension to ordinary
differential equations which allows modelling of noisy and uncertain driving
forces. These models are particularly appealing due to their flexibility in
expressing highly complex relationships with simple equations, while
retaining a high degree of interpretability.
Much work has been done over the last century focussing on understanding and
modelling with SDEs, particularly in dynamical systems and quantitative
finance~\citep{StochasticProcesses, Malliavin}.
Examples of such work are the Langevin model of stochastic dynamics of particles
in a fluid, stochastic turbulence~\citep{StochasticTurbulence}, and the
Black--Scholes model. More recently, SDE models have gathered interest in
machine learning and current work focuses on efficient and scalable methods of
inferring SDEs from data.

\subsection{The SDE learning problem}

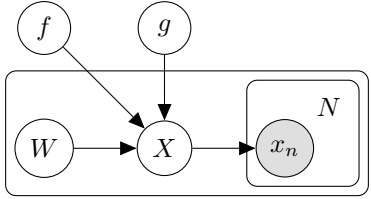
\begin{figure}[htb]
  \begin{center}
    \begin{tikzpicture}
      \tikzstyle{rv} = [
      circle,fill=white,draw=black,node distance=1.6cm,minimum size=0.7cm
      ]
      \tikzstyle{plate} = [draw, rectangle, rounded corners, fit=#1]

      \node[rv] (f) {$f$};
      \node[rv, right of=f] (g) {$g$};
      \node[rv, below of=f] (W) {$W$};
      \node[rv,below of=g] (X) {$X$};
      \node[rv,fill=gray!25,right of=X] (x) {$x_n$};

      \path (W) edge [->, >={triangle 45}] (X);
      \path (f) edge [->, >={triangle 45}] (X);
      \path (g) edge [->, >={triangle 45}] (X);
      \path (X) edge [->, >={triangle 45}] (x);

      \node[above right=1pt of x] (N){$N$};
      \node[plate=(x)(N)] (plate-T) {};
      \node[plate=(W)(X)(plate-T)] {};
    \end{tikzpicture}
  \end{center}
  \caption{\label{fig:pgm}Graphical representation of an SDE model.}
\end{figure}

Consider an It\^o SDE~\citep{StochasticProcesses} of the following form
\begin{equation}
  \label{eq:sde}
  \dd{X} = f(X, t) \dd{t} + g(X, t) \,\dd W,
\end{equation}
and a set of observations $x_{1:N} = \bigl\{(t_n, x_n)\bigr\}_{n=1}^N$. In the
simplest case, these observations would be directly of a realisation of the
process at discrete points in time. In many physical applications, the
observations themselves would have some uncertainty associated with them,
described by an observation noise model. In more complex models, the SDE would
be driving some unobserved parameters describing further complex processes. For
example, discrete events may be modeled by an inhomogenious Poisson processes,
where the rate parameter is evolving stochastically over time. In any case, the
generating process can be summarized by the graphical model in
Figure~\ref{fig:pgm} and the corresponding learning problem can be described
as: ``given observations $x_{1:N}$, infer $f$ and $g$''.

For the most part, the model can be trained using standard techniques, such as
variational inference, Markov chain Monte Carlo or optimizing an appropriate
loss function. Generally, this requires estimating an expectation over
sample-paths, either to estimate the marginal probability of the observations as
a function of $f$ and $g$, or to estimate the mismatch of sample-paths with
the observations.
To estimate these expectations, most current work makes use of integration
algorithms. These algorithms generate trajectories of $X$ for a given $f$ and
$g$, which can be used to estimate the required expectation. For example, the
Euler--Maruyama algorithm~\citep{NeuralSDE, NeuralJumpSDE, StabilizingNeuralODE}
generates trajectories with the following update rule
\begin{equation}
  \label{eq:euler}
  X_{t+\Delta t} \gets X_{t} + f(X_{t},t) \, \Delta t + g(X_{t}, t) \, \Norm{0, \Delta t}.
\end{equation}
The expectation of some functional $F$ is estimated using the sample mean of its
value over many independent trajectories $X^{(k)}$
\begin{equation}
\hat{F} := \frac{1}{K} \sum_{k=1}^{K} F[X^{(k)}].
\end{equation}
 
Unfortunately, integrators such as Equation~\eqref{eq:euler} introduce a sequential
dependence between each point and the next, which inhibits parallelization.
Furthermore, it introduces difficulties in probabilistic modelling of the
functions $f$ and $g$; in order to properly estimate the expectation, the sample
paths should be generated under consistent realisations of $f$ and $g$. Models
which implicitly describe $f$ as a Gaussian process, would need to sample
$f(X_{t}, t)$ conditioned on all the previously sampled values of $f$ along the
trajectory, which may be computationally prohibitive.

The integrator generates trajectories sampled from the prior distribution over
paths. In the initial stages of learning, the path-space prior distribution
usually has minimal overlap with the path-space posterior and typical
trajectories do not pass anywhere near the observations, resulting in high
variance gradients. This grows considerably worse as the dimension of the
problem and the time between observations increases.

The central idea of this paper is to instead generate trajectories from a
distribution much closer to the path-space posterior distribution when
estimating these expectations. In the simplest case where the observations are
directly of the process $X$, this means sampling trajectories which pass
directly through the observations. Alternatively, one can marginalize out the
observation noise distribution --- or efficiently estimate some expected loss
function --- by importance sampling
\begin{equation}
  \E_X\bigl[p(x_{1:N} \,|\, X)\bigr] =
  \E_{\tilde{x}_{1:N}\sim q}\left[
    \frac{p(x_{1:N} \,|\, \tilde{x}_{1:N})}{q(\tilde{x}_{1:N})} p_X(\tilde{x}_{1:N})
  \right],
\end{equation}
where $q$ is an importance distribution which may depend on the observations
themselves. For a large class of observation distributions, such as additive
Gaussian noise or multinomial observations, a natural choice for $q$ would be
the corresponding conjugate distribution, in which case the importance weight
$p/q$ becomes a constant. In complex graphical models, $q$ might be
parameterized and optimized variationally, or Gibbs sampling may be used with a
collection of simpler conditional distributions.
Taking advantage of this, we shall, without loss of generality, concentrate only
on estimating the quantity
$p_X(x_{1:N}) = \E_{X}\Big[\prod_n\delta\bigl(x_n-X(t_n)\bigr)\Big]$.

Our contributions are as follows:
\begin{itemize}
\item We introduce a novel algorithm for efficient estimation of probabilities
  in SDE models.
\item Our algorithm exhibits lower gradient variance in estimated expectations
  than algorithms based on SDE solvers, since it enforces that sampled
  trajectories pass through the observed data, producing accurate results even
  in hundreds of dimensions.
\item Since our algorithm does not rely on an SDE solver, it completely removes
  all sequential dependence between sampled points. Our algorithm is therefore
  trivially parallelizable, and can take full advantage of modern parallel
  hardware such as multi-GPU training.
\item This approach enables inference for models with or without an explicit
  observation noise model. This is in contrast to integrator-based methods, which
  must add observation noise to account for the mismatch between observations and
  sample paths, even when it is not required by the model.
\item Since our estimator can be calculated in a single forward pass
  of the function $f$, probabilistic approaches such as variational Gaussian
  processes are viable candidates for the representation of $f$.
\end{itemize}

\subsection{Related work}

\citet{NeuralSDE} propose a method for calculating gradients of functions of the
integrated process with respect to parameters of $f$ and $g$ via a backwards SDE
called the adjoint method. This approach requires reconstructing the Wiener
process backwards in time when integrating the backward equation and the
authors propose a memory efficient algorithm to do so. Their approach uses a
constant amount of memory independent of the number of integration steps similar
to the adjoint method for NeuralODEs~\citep{NeuralODE}.
\citet{NSDELatentGaussian} discuss the computation of gradients by simulation of
a forward SDE in the derivatives with respect to parameters of the model using
the chain rule of stochastic calculus. This approach does not require
back-propagating through the SDE solver. \citet{SDEGAN} propose a method of
learning SDEs by treating the SDE as the generator in a generative adversarial
network setup, using a second SDE for the discriminator. All of these
approaches use SDE integration algorithms to estimate path-space expectations.

\citet{ApproxBayesSDE} consider a nonparametric approach to SDE learning, using
Gaussian processes for the drift and diffusion functions. They initially
consider a gradient matching approximation, in which the drift Gaussian process
may be fit directly by conventional methods, and thereafter use an
expectation-maximization algorithm, and estimate the transition probability for
sparse observations using a linear approximation of the process.
\citet{LearningStochasticDifferential} propose an alternative approach to
Gaussian process-based SDE learning using the Euler--Maruyama integrator,
approximating $f$ by the predictive mean of a Gaussian process conditioned on a
MAP estimate for a set of inducing points. Unfortunately, this approach
completely ignores any uncertainty in the posterior over $f$.

Much other work focuses on learning specific classes of SDE or ODE models, such
as symplectic or Hamiltonian
systems~\citep{SymplecticOdeNet,HamiltonianNeuralNetworks}, graph-based
models~\citep{GraphNeuralOrdinary,HamiltonianGraphNetworks}, and controlled
differential equations~\citep{NeuralCDE}. Many of these ideas are independent of
the learning algorithm and can potentially make use of our algorithm when
applied to SDEs.

\section{Path-space importance sampling}

It is well known that solutions of linear SDEs are Gaussian
processes~\cite[see][Section~3.7]{StochasticProcesses}.
For these processes, the joint probability density over
some finite collection of points is available in closed form. Furthermore, the
posterior distribution of the process conditioned on some set of observations
can be sampled from exactly.
Unfortunately, this is not the case in general for non-linear SDEs, and one has
to resort to approximation methods in order to calculate probabilities.
Most commonly, one uses an SDE integrator to estimate expectations; however,
other methods, such as Laplace approximations and perturbation theory are
sometimes used \citep{LaplaceSDE,LaplaceBasedApproximate,SDEPerturbationTheory}.

Alternatively, if we are able to find a second process which follows a similar
SDE to the one of interest, it is possible to express expectations and
probabilities as importance sampling estimators.
Linear SDEs are an attractive candidate for this due to their Gaussianity and
ease of sampling.

\subsection{State-independent diffusion}

For now, consider the simpler scenario in which the diffusion coefficient does not
depend on the state variable; i.e.\ $g(x, t) = \sigma(t)$ for some function
$\sigma$.
The process defined by
\begin{equation}
\label{}
\dd{Y} = \sigma(t) \,\dd W,
\end{equation}
is Gaussian with mean zero and conditional variance
\begin{equation}
\label{}
\E_Y\left[ \bigl( Y(t_1) - Y(t_0) \bigr)^2 \right] = \int_{t_0}^{t_1} \sigma(t)\sigma^T(t) \dd{t}.
\end{equation}

Sample paths of $Y$ can be generated efficiently by simulating Brownian motion.
The process $Y$ conditioned on a set of observations of the process is a
Brownian bridge.
If we are able to express quantities of interest as expectations under this
process instead of our original SDE, then we can efficiently estimate such
quantities via Monte Carlo sampling.
As it turns out, expectations and probabilities for an SDE with
a general drift term can in fact be expressed as such.
This relation is given in the following theorem.

\begin{theorem}
  \label{th:rn-deriv}
  Let $X$ and $Y$ be the stochastic processes generated by the following SDEs
  \begin{align}
    \label{}
    \dd{X} &= f(X, t) \dd{t} + \sigma(t) \,\dd W, \\
    \dd{Y} &= \sigma(t) \,\dd W.
  \end{align}
  Further assume that $\sigma(t)\sigma^T(t)$ is Lebesgue-almost-everywhere
  invertible
  and $f$ is sufficiently well-behaved such that $X(t)$ has a unique
  probability density for all $t$.

  The probability density of observations $\bigl\{ (t_n, x_n) \bigr\}_{n=1}^N$,
  under the process $X$ is given by the conditional expectation
  \begin{equation}
    p_X(x_{1:N}) = p_Y(x_{1:N}) \E_Y \left[ e^{S[Y]} \,\middle|\, \bigl\{ Y(t_n) = x_n \bigr\}_{n=1}^N \right],
  \end{equation}
  where
  \begin{equation}
    \label{eq:girsanov}
    S[Y] = \int f^T(Y_t, t) \bigl(\sigma(t)\sigma^T(t)\bigr)^{-1} \dd{Y_t}
    - \frac{1}{2} \int f^T(Y_t, t) \bigl(\sigma(t)\sigma^T(t)\bigr)^{-1} f(Y_t, t) \dd{t}.
  \end{equation}
\end{theorem}
This result follows from Girsanov's theorem~\citetext{\citealp{Girsanov};
\citealp[Section~1.5]{Malliavin}} and the definition of conditional expectation.
Intuitively speaking, the first term in Equation~\ref{eq:girsanov} encourages
$f$ to line up with typical paths which pass through the observations, while the
second term regularizes the magnitude of $f$.

Theorem~\ref{th:rn-deriv} allows us to develop an importance sampling algorithm
for estimating probabilities under the process $X$ by simulating the process
$Y$, where the importance weights are given by $\omega_i = \mathrm{e}^{S[Y^{(i)}]}$.
In this approach, one can generate entire sample paths of $Y$ before calculating
$f(Y, t)$, allowing the forward pass of $f$ to be performed in parallel.
This approach is described in Algorithm~\ref{alg:diff-ind}.

\begin{algorithm}[htb]
  \caption{Path Integral Importance Sampling}
  \label{alg:diff-ind}
  \begin{algorithmic}[1]
    \For{$i = 1,\ldots, K$}
    \State Sample $(Y_t)_{t=t_1}^{t_N}$, s.t. for each $n$, $Y_{t_n} = x_n$
    \label{alg:sample}
    \Comment{$Y$ is a Brownian bridge}
    \State $f_t \gets f(Y_t, t)$ for each $t$
    \State $\alpha \gets \sum_t f_t^T \sigma^{-2}(t) \, (Y_{t+\Delta t} - Y_t)$
    \State $\beta \gets  \sum_t f_t^T \sigma^{-2}(t) f_t \,
    \Delta t$
    \State $S_i \gets \alpha - \frac{1}{2} \beta$
    \EndFor
    \State \Return $p_Y(x_{1:N}) \, \frac{1}{K} \sum_i \exp(S_i)$
  \end{algorithmic}
\end{algorithm}
\begin{figure}[htb]
  \centering
  \begin{subfigure}{0.49\textwidth}
    \includegraphics[width=\textwidth]{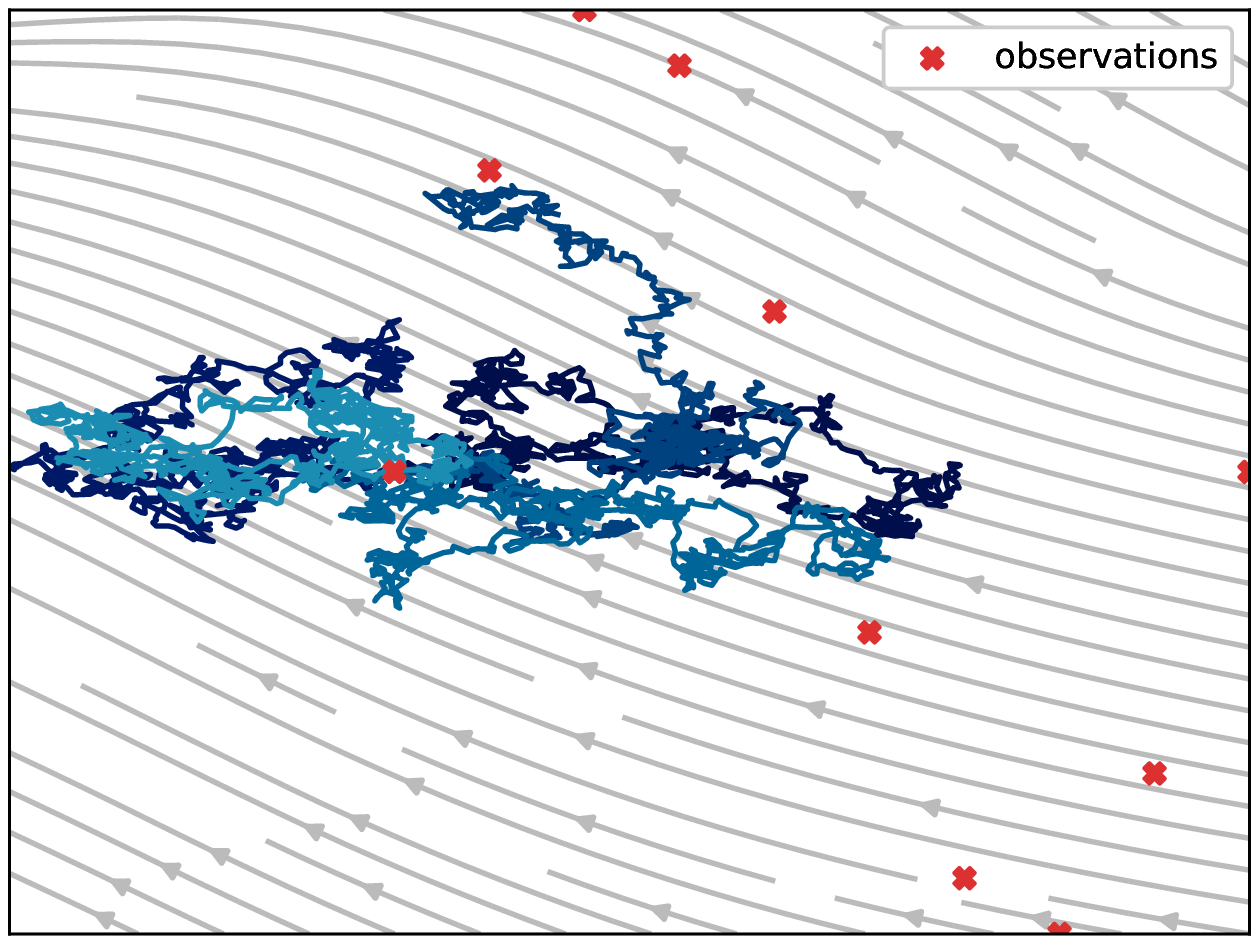}
    \subcaption{\label{fig:sdeint}SDE integration}
  \end{subfigure}
  ~
  \begin{subfigure}{0.49\textwidth}
    \includegraphics[width=\textwidth]{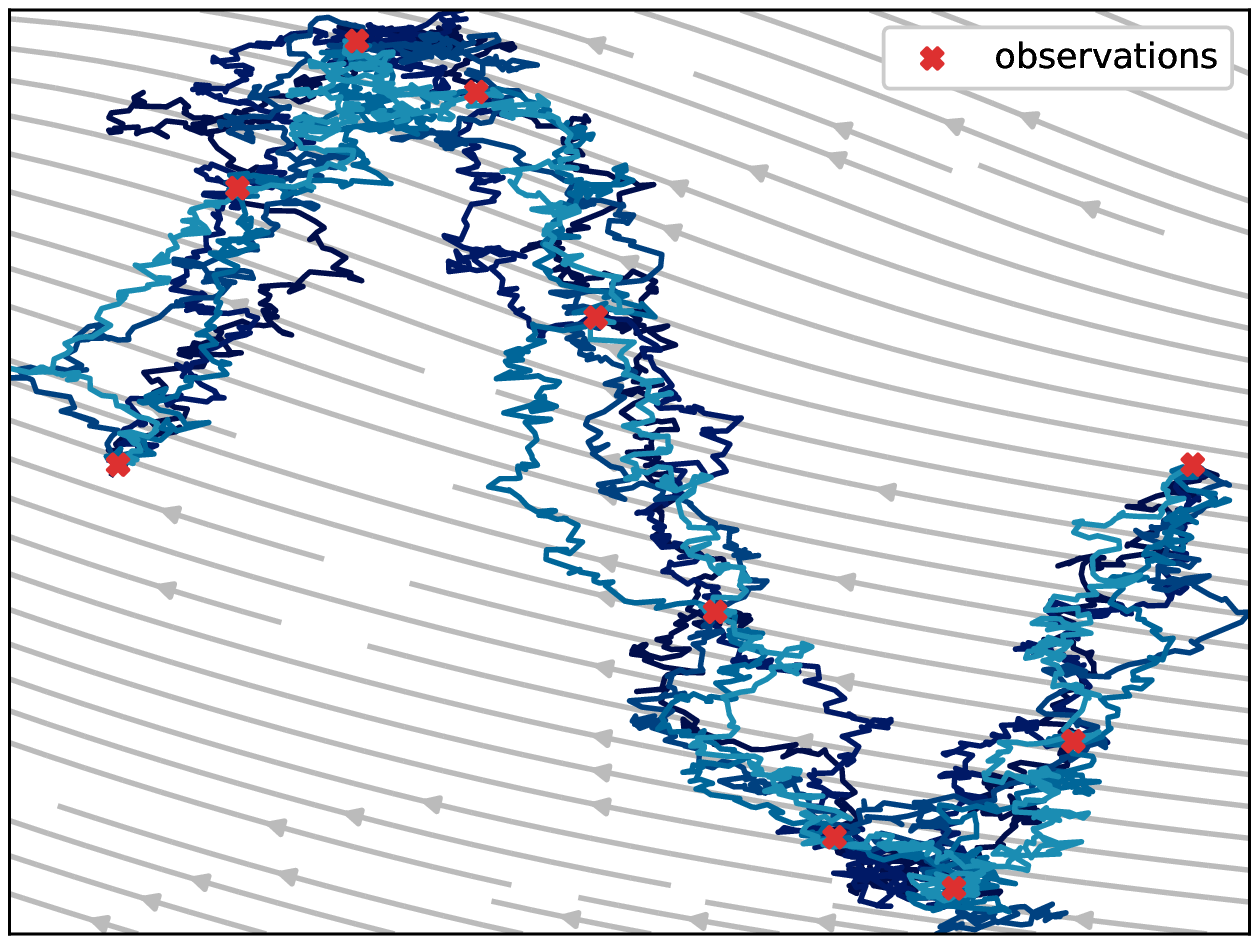}
    \subcaption{\label{fig:bridge}Brownian bridge importance samples}
  \end{subfigure}
  \caption{\label{fig:samples} Independently sampled paths.
    \subref{fig:sdeint} shows paths sampled from the SDE using an integrator,
    while \subref{fig:bridge} shows samples from a Brownian bridge which pass
    exactly through the observations. In both cases the drift function is given
    by a neural network which is randomly initialized with the same seed. The
    drift vector field is shown in grey.}
\end{figure}

An important property of this algorithm is that the sampled paths directly pass
through observations. This means that the probability estimates directly relate
the drift function $f$ to plausible trajectories instead of allowing the
particles to wander around randomly. See Figure~\ref{fig:samples} for an
illustration.
In the initial stages of training, simply integrating the SDE with drift $f$ will
typically not lead to trajectories which pass near the data, especially in high
dimensions. This can have a significant impact on the variance of the gradients
and can inhibit the model's ability to learn.
Training on paths that pass through the data can greatly mitigate this high
variance.
As with other algorithms which generate SDE sample paths, the discretization of
the integrals introduces a small bias. However, this bias vanishes in the
$\Delta t \to 0$ limit.

\subsubsection{A note on parallelizability}

With the exception of the sampling step in Line~\ref{alg:sample} of
Algorithm~\ref{alg:diff-ind}, each of the other operations, including the sums,
can be performed in parallel over feature and batch dimensions, as well as over
the $t$ and $i$ indices. Sampling a Brownian bridge, or any conditional
Gauss--Markov process, can be performed by first sampling the unconditional
process, and then linearly interpolating with the observations. The linear
interpolation can be performed independently, and therefore in parallel, over
the time dimension. The only operation that is not completely parallelizable is
the sampling of the unconditional base process --- in this case standard
Brownian motion. Brownian paths can be sampled by cumulative summation of
independent normal random numbers. However, this cumulative sum is extremely
fast, and can be performed independently between each consecutive pair of
observations if need be. In our tests, sampling whole Brownian trajectories was
about 5 orders of magnitude faster than the forward pass of the neural network
$f$, and so does not create any noticeable performance impact.

Once the Brownian paths are sampled, the rest of Algorithm~\ref{alg:diff-ind}
may be implemented in parallel without difficulty. Contrast this to standard
integrator-base methods; even if the Brownian paths are sampled upfront, the
integrator must still perform each of the forward passes of $f$ in sequence.

See
Appendix~\ref{app:bridge} for more details on sampling Gaussian bridges.

\subsection{State-dependent diffusion}
\label{sec:state-dependent}

The assumption that $g$ is independent of the state is not in fact required for the
validity of Theorem~\ref{th:rn-deriv};
the reason this assumption is required is to ensure that the process $Y$ is
Gaussian, so as to enable easy conditional sampling and calculation of
$p_Y(x_{1:N})$.
Unfortunately, one cannot simply find an analogue of Theorem~\ref{th:rn-deriv}
to calculate expectations of a general process as expectations with respect to a
constant diffusion process. 

To address the question of state-dependent diffusion, it is enlightening to
consider a simple example of such a process. Perhaps the most common example is
geometric Brownian motion
\begin{equation}
\label{}
\dd{X} = \mu X \dd{t} + \sigma X \,\dd W.
\end{equation}
The typical way in which one would solve this is to introduce a transformation
$Z = \log(X)$, and apply It\^o's
lemma~\citetext{\citealp[see][Section~1.5]{Malliavin};
\citealp[Section~3.5]{StochasticProcesses}} to obtain the SDE
\begin{equation}
\label{}
\dd{Z} = \bigl( \mu - \tfrac{1}{2}\sigma^2 \bigr) \dd{t} + \sigma \,\dd W.
\end{equation}

Similarly, one might address a problem of a general state-dependent diffusion
coefficient by considering a transformation to a space in which the transformed
equation has constant diffusion.
Let $T(\cdot, t)$ be a map which is invertible for all $t$. That is, there
exists a function $T^{-1}$ such that if $y = T(x, t)$, then $x = T^{-1}(y, t)$.
Transforming a process with constant diffusion $\sigma$, by the map $T^{-1}$ gives a
process with diffusion coefficient
\begin{equation}
\label{}
g_{i,j}(x, t) = \frac{\partial T^{-1}_{i}}{\partial y_k} \sigma_{k,j}
= \left( \frac{\partial T(x, t)}{\partial x} \right)^{-1}_{i,k} \sigma_{k,j},
\end{equation}
where repeated indices imply summation.
With such a transformation we can represent any diffusion matrix which can be
written as a Jacobian.
While the restrictions on the diffusion matrix may be seen as an inherent
limitation of a path-space importance sampling approach, it may be argued that
constant or simple diffusion functions would cover the majority of cases of
interest, and almost all real-world problems require some form of data
pre-processing and, potentially, link functions.

This transformation allows us to infer the SDE in a transformed space, with
constant diffusion, and, if desired, transform the equation back to the original
space using It\^o's lemma. 
This approach is described in Algorithm~\ref{alg:diff-dep}, where we have used
the symbol $\tilde f$ to represent the drift in the transformed space as to
avoid ambiguity with the drift in the original space.
One important point to note is that one should be careful to allow gradients to
propagate through the sampling step on Line~\ref{alg:dep:line:sample} in
Algorithm~\ref{alg:diff-dep}, so as to allow learning of the function $T$.
\begin{algorithm}[htb]
  \caption{Transformed-State Path Integral Importance Sampling}
  \label{alg:diff-dep}
  \begin{algorithmic}[1]
    \For{$i = 1,\ldots, K$}
      \State Sample $(Y_t)_{t=t_1}^{t_N}$, s.t. for each $n$, $Y_{t_n} = T(x_n, t_n)$
      \label{alg:dep:line:sample}
      \Comment{$Y$ is a Brownian bridge}
      \State $\tilde f_t \gets \tilde f(Y_t)$ for each $t$
      \State $\alpha \gets \sum_t \tilde f_t^T \sigma^{-2} \, (Y_{t+\Delta t} - Y_t)$
      \State $\beta \gets  \sum_t \tilde f_t^T \sigma^{-2} \tilde f_t \,
      \Delta t$
      \State $S_i \gets \alpha - \frac{1}{2} \beta$
    \EndFor
    \State \Return $\prod_n \det\left( \frac{\partial T(x_n, t_n)}{\partial x_n} \right) p_Y\bigl( T(x_{1:N},t_{1:N}) \bigr) \, \frac{1}{K} \sum_i \exp(S_i)$
  \end{algorithmic}
\end{algorithm}

One would typically apply this method to infer $\tilde f$ rather than the
original $f$ for performance reasons, and then only reconstruct the SDE in the
original space if need be.

\subsubsection{Reconstruction of the SDE in the untransformed space}

Generation of new sample paths can be performed easily in the transformed space
and then simply transforming back by applying $T^{-1}$.
However, sometimes one needs more information for the purposes of analysing the
learned equation.
In this case the SDE can be transformed back using It\^o's lemma. The drift and
diffusion coefficients are

\begin{align}
\label{eq:ito-drift}
f_i(x, t)
  &= 
    \frac{\partial T^{-1}_i}{\partial t}
    + \frac{\partial T^{-1}_i}{\partial y_k}\tilde f_k(y, t) 
    \:+\: \frac{1}{2} \sigma_{j,k} \frac{\partial^2 T^{-1}_i}{\partial y_j\partial y_l} \sigma_{l, k},
  \\
\label{eq:ito-diff}
g_{i,j}(x, t)
  &= \frac{\partial T^{-1}_i}{\partial y_k} \sigma_{k,j}.
\end{align}
All expression are evaluated at $y = T(x, t)$.
When $T$ is parameterized by a neural network with ReLU non-linearities, the
second derivative term in Equation~\eqref{eq:ito-drift} is almost-everywhere
zero. The remaining terms are Jacobian vector products, and so can be performed
efficiently even in high dimensions.\footnote{However, this may require
  forward-mode instead of backward-mode differentiation.}

\section{Experiments}

In this section, we provide some experiments to illustrate and verify the
performance and capabilities of our proposed algorithm.
Further details on the experiments are given in Appendix~\ref{app:exp}.

\subsection{The Lorenz system}
\label{sec:lorenz}

To compare the effect of our algorithm on learning, we follow~\citet{NeuralSDE}
and train a neural network based SDE model on the Lorenz system. In this model
the drift function is given by a multi-layer perceptron, and the diffusion term is a
simple scalar.
We generated 16 independent paths, consisting of $200$ observations each,
uniformly sampled over a time interval of $[0, 10]$.
We trained the model on this data by two different methods. In the first method,
we minimize the mean-squared error, estimated by integrating out the SDE, and
calculating gradients using the adjoint
\mbox{method~\citep{NeuralSDE}}.\footnote{We also attempted training a model
  without the adjoint method and instead directly backpropagating through the
  SDE integrator. However, a single forward pass required more memory than
  available on the GPU.}
In the second method, we directly optimize the log-probability calculated using
Algorithm~\ref{alg:diff-ind}.
In both cases we used precisely the same optimization parameters, the same
time-step size $\Delta t$, and the same number of sampled trajectories to
calculate averages.
Both models used the same seed for the neural network initialization, and the
same number of gradient steps were performed in both cases.
These models were trained on an Nvidia~RTX~3070, with 8Gb of memory.

\begin{figure}[ht]
  \centering
  \begin{subfigure}{0.49\textwidth}
    \includegraphics[width=\textwidth]{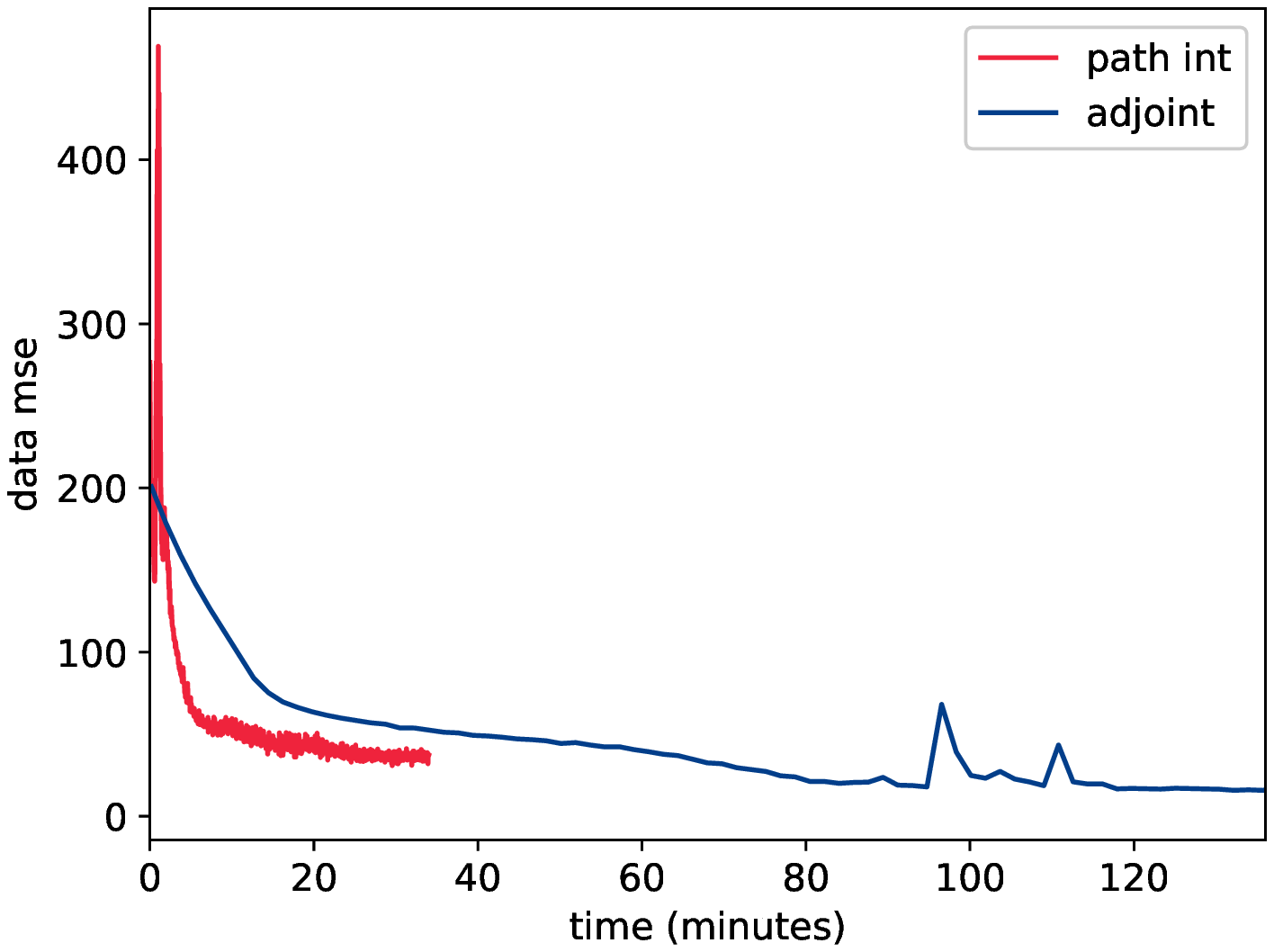}
    \subcaption{\label{fig:mse}}
  \end{subfigure}
  \begin{subfigure}{0.49\textwidth}
    \includegraphics[width=\textwidth]{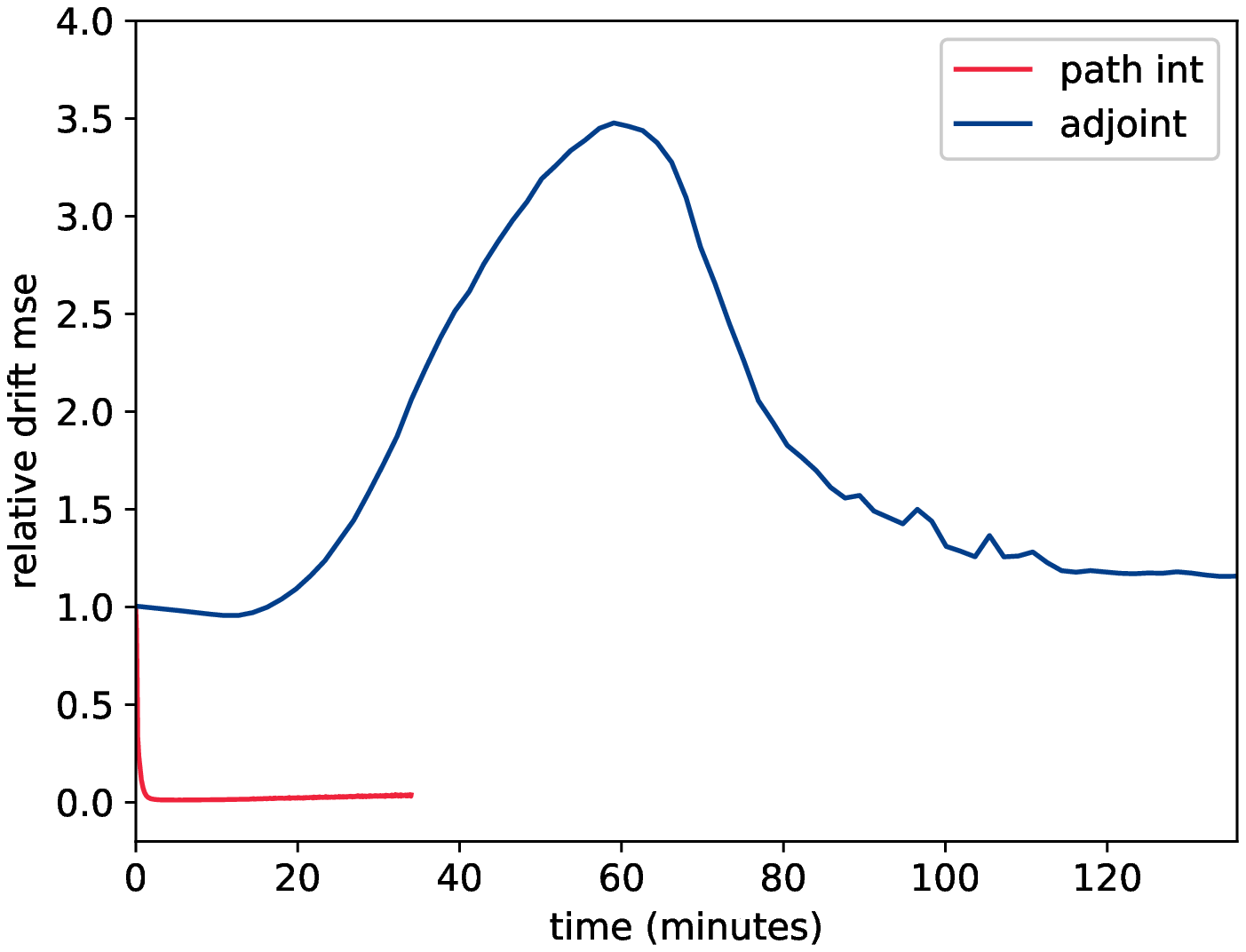}
    \subcaption{\label{fig:drift}}
  \end{subfigure}
  \caption{\label{fig:lorenz}Learning curves of a neural SDE on the Lorenz
    system.
  \subref{fig:drift} shows the normalized mean-squared deviation of the learned
  drift from the ground truth. \subref{fig:mse} shows the mean-squared error of
  sample paths from the observations.}
\end{figure}

To assess the performance of these approaches, we record two metrics.
The first is the mean-square difference between the learned drift vector field,
and the Lorenz vector field, normalized by the mean-square magnitude of the
Lorenz vector field, measured at the observations.
Written mathematically:
\begin{equation}
  \operatorname{err} = \frac{
    \sum_n \bigl(
    f_\theta(x_n) - f_{\mathrm{Lor}}(x_n)
    \bigr)^2
  }{
    \sum_n \bigl(
    f_{\mathrm{Lor}}(x_n)
    \bigr)^2
  }
\end{equation}
This metric is shown in Figure~\ref{fig:drift}.
The second metric we record the mean-squared error of sampled trajectories
compared to the observations; this is shown in Figure~\ref{fig:mse}. However, we
note that this is not a fair performance metric since the adjoint-based method
is directly optimizing for it.
On the x-axis we show the training time in minutes.

In Figure~\ref{fig:lorenz} we have limited the range of the x-axis in order for
the lines to be more clearly visible. The total training time for the
adjoint-based MSE optimization was a factor 52 longer than the time taken
optimizing the probability estimates.
We note that, while our approach was significantly faster, it did use
approximately four times as much memory as the adjoint-based
method.\footnote{The adjoint algorithm was specifically designed to have
  constant memory complexity in the number of integration steps. Our algorithm
  does not exhibit this feature.}

The poor performance of the integrator-based learning algorithm in
Figure~\ref{fig:drift} --- despite its low mean-squared-error --- suggests that
directly optimizing for the discrepancy of typical SDE sample-paths compared to
observations is not a robust approach to SDE learning.

\subsection{Gradient variance}

\begin{figure}[htb]
  \centering
  \begin{subfigure}{0.49\textwidth}
    \includegraphics[width=\textwidth]{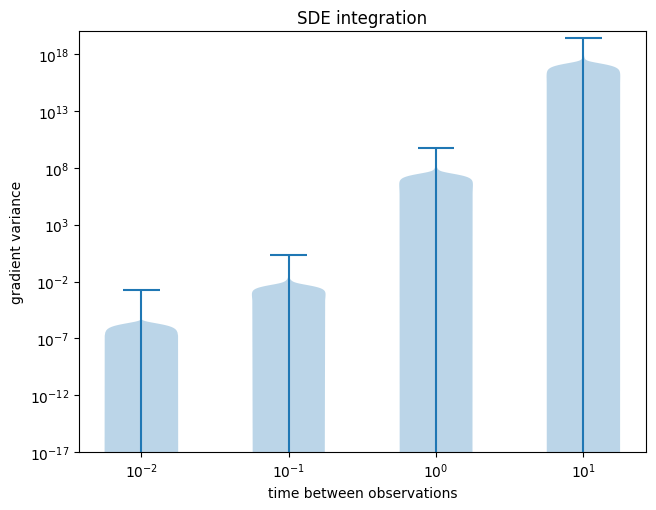}
    \subcaption{\label{fig:grad-var-adj}}
  \end{subfigure}
  \begin{subfigure}{0.49\textwidth}
    \includegraphics[width=\textwidth]{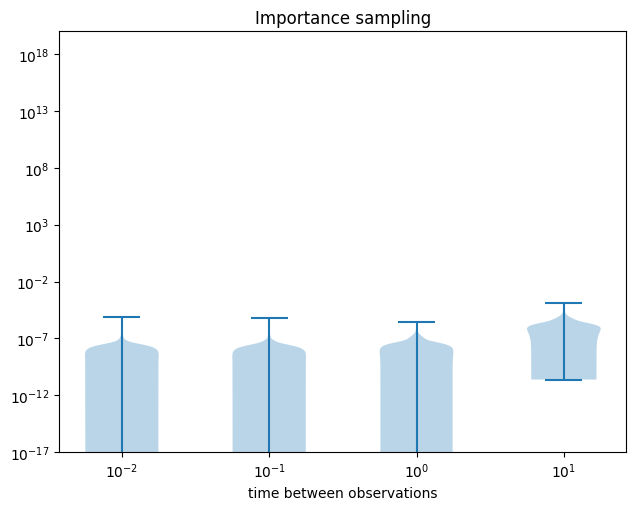}
    \subcaption{\label{fig:grad-var-pathint}}
  \end{subfigure}
  \caption{\label{fig:grad-var}Gradient variances as a function of the time
    between observations. \subref{fig:grad-var-adj} shows gradient variances of
    the observation mean-squared-error computed with an SDE integrator.
    \subref{fig:grad-var-pathint} shows the gradient variance of the
    log-probability of observations computed with Algorithm~\ref{alg:diff-ind}.
  }
\end{figure}

We applied both our algorithm and an SDE integrator to a neural network-based
SDE model on a real-world dataset and recorded the gradients, sampled over many
independent trajectories. We thereafter computed the variance of the gradients
individually per parameter, for various time lengths between observations. We
used a discretization step size of 0.01. These results are shown in
Figure~\ref{fig:grad-var}.

In each case we used the same neural network, initialized with a fixed seed, and
the exact same values for all parameters. We used the same loss functions as in
Section~\ref{sec:lorenz}. The data set we used was the
Hungarian chickenpox cases dataset from the UCI machine learning
repository;\footnote{
  This dataset can be found at
  \url{https://archive.ics.uci.edu/ml/datasets/Hungarian+Chickenpox+Cases}
}
it has 20 features per observation. We scaled the features by their standard
deviations before performing the experiment to improve numerical stability.

Our findings are as follows:
\begin{itemize}
\item 
  For very short times, we found that both algorithms produced comparable
  gradient variances: typically less than $10^{-7}$.
\item 
  For longer time spans (3 orders of
  magnitude larger than the discretization step size), the gradient variance of
  our algorithm increased by a few orders of magnitude; the maximum we observed
  was on the order of $10^{-4}$. However, the SDE integrator approach resulted in
  huge increases in variances, with some as large as $10^{18}$.
\item 
  Between the above
  two extremes, we found that the gradient variance of our approach did increase
  with the time between observations, but stayed below $10^{-5}$, while the gradient
  variance from the SDE integrator exhibited very poor scaling with variances of
  up to $10^{8}$ for times between observations of 1.0.
\end{itemize}

Considering that both approaches produced comparable variances on very small
timescales, we feel that this experiment represents a fair comparison.

\subsection{Uncertainty quantification with variational Gaussian processes}

We attempt to test the uncertainty quantification capabilities of using a GP SDE
model, and the viability of using our method with such a model. To assess this,
we used data trajectories from a van der Pol oscillator. For the GP model we
used a grid-based inducing point approach and the kernel interpolation method of
\citet{KissGP}. We used a zero mean function and a radial basis function
kernel, with independent GP priors over each output dimension of the drift
vector field. The combination of these techniques allows one to take advantage
of the Kronecker product structure of the covariance matrices and the Toeplitz
structure of each factor for accelerated sampling.

\begin{figure}[htb]
  \centering
  \begin{subfigure}{0.32\textwidth}
    \includegraphics[width=\textwidth]{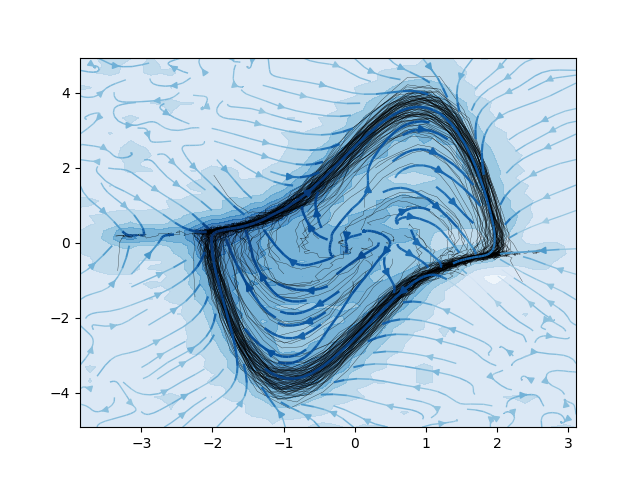}
    \caption{\label{fig:gp-post}}
  \end{subfigure}
  \begin{subfigure}{0.32\textwidth}
    \includegraphics[width=\textwidth]{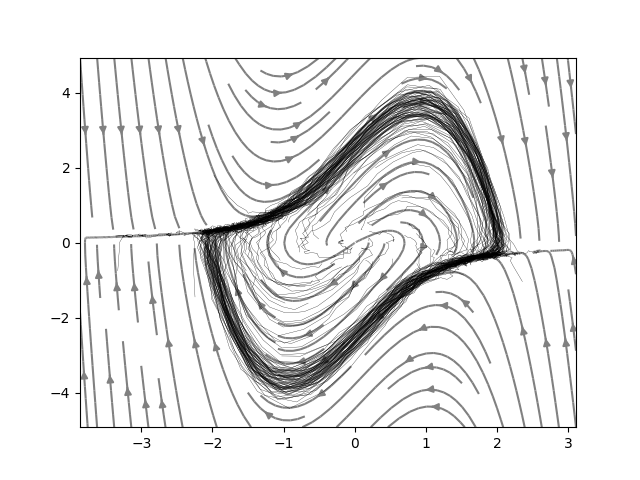}
    \caption{\label{fig:vdp}}
  \end{subfigure}
  \begin{subfigure}{0.32\textwidth}
    \includegraphics[width=\textwidth]{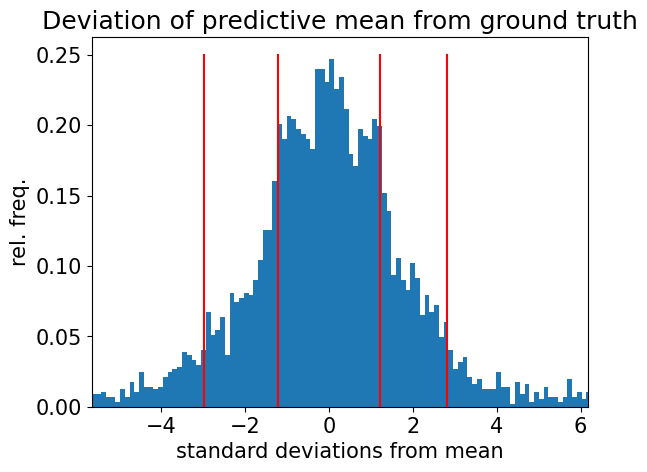}
    \caption{\label{fig:gp-err}}
  \end{subfigure}
  \caption{
    \label{fig:var-gp}GP posterior compared to ground truth vector field.
    \subref{fig:gp-post} shows the GP vector field coloured by uncertainty. The
    arrows show the integral curves of the posterior mean and the coloured are
    scaled linearly with the log predictive variance. \subref{fig:vdp} shows the
    integral curves of the ground truth van der Pol oscillator.
    \subref{fig:gp-err} shows the deviation of GP posterior from the ground
    truth vector field. The histogram shows the difference of the ground truth
    and predictive mean measured in standard deviations. The red lines show the
    10\%, 25\%, 75\%, and 90\% quantiles.
  }
\end{figure}

The approximate GP posterior is shown in Figure~\ref{fig:gp-post} compared to
the ground truth vector field of the van der Pol oscillator in Figure~\ref{fig:vdp}.
Comparing the GP to the van der Pol vector field, we found that the GP model was
fairly accurate in areas with many data points, while reverting to the prior far
away from observations. To assess the quality of the uncertainty quantification,
we further calculated the discrepancy between the variational posterior and the
ground truth vector field. Over a dense grid, covering an enlarged area around
the observations, we calculated the discrepancy as
\begin{equation}
  \delta=\frac{f_{\mathrm{vdp}}(x) - \mu(x)}{\sigma(x)}.
\end{equation}
A histogram is shown in Figure~\ref{fig:gp-err}.

Under Gaussianity assumptions, one expects to see the quantiles line up well
with some associated number of standard deviations. We found that the
distribution of these $\delta$s was symmetric, but it was fatter-tailed than a
Gaussian distribution. We found that the middle 50\% (25\% - 75\% quantiles)
slightly passed one standard deviation on either side, as opposed to the 68\% of
a standard Gaussian distribution. The middle 80\% (10\% - 90\% quantiles) were
within 3 standard deviations. The van der Pol vector field grows polynomially
away from the origin, which is likely the main reason for this discrepancy, and
may indicate that a more flexible kernel function is required, such as a
combination of squared exponential and polynomial kernels.
While the posterior uncertainty was
somewhat overconfident (as is known to happen with variational inference), it
was not wholly unreasonable.

We further note that, while the
complexities involved in variational GP fitting can result in longer training
times, we still found that our method applied to a GP was significantly faster
than using an SDE integrator with a neural network.

\section{Discussion}

In this work we introduced an algorithm to efficiently estimate probabilities of
observations under the assumption of an SDE model for the purposes of learning.
The proposed approach produces estimates which are accurate, even in high
dimensions, and can fully utilize parallel hardware for significant speedups.

The proposed algorithm is capable of providing robust probability estimates which can be
used in models both with or without an explicit observation noise component.
This is in contrast to SDE integrator-based learning approaches, which must
incorporate and estimate an explicit error term due to their inability to generate paths which are
guaranteed to pass through the observations.

Our algorithm produces more stable estimates than alternative integrator-based approaches,
particularly when the observations are sparsely distributed over long time
intervals and in high dimensions. This in turn results in more stable and robust
learning. With the rise in applicability of large-scale (neural network) SDE modeling in
science, simulation, and domains such as finance, we see our approach as
an important contribution to reliable scaling of these methods.


\ificlrfinal
\subsubsection*{Acknowledgments}

The authors would like to thank InstaDeep Ltd.\ for their financial support of this work.
\fi

\subsubsection*{Reproducibility}

Efforts have been made to describe all parameters and details of the algorithms
used in this paper, either in the content or in the appendix. The code included
in the supplementary material includes a Dockerfile and instructions for running
some of the experiments. Where relevant, fixed seeds for random number
generators can be found in the code during initialization.

\bibliography{paper.bbl}
\bibliographystyle{iclr2022_conference}

\appendix
\section{Theorem~\ref{th:rn-deriv}}

We have that $X$ and $Y$ are processes obeying the SDEs
\begin{align}
  \dd{X} &= f(X) \dd{t} + g(X) \,\dd W, \\
  \dd{Y} &= g(Y) \, \dd W,
\end{align}
dropping the time argument for notational simplicity.
Expectations with respect to SDEs can be expressed as path integrals.
In the It\^o convention, the path integral representation of an SDE has the
following form
\begin{align}
  \E_{X}\bigl[ F[X] \bigr]
  &= \lim_{N\to\infty} \int F[x]
    \prod_{n=1}^N \frac{\dd{x_n}}{\sqrt{2\pi\Delta t_n} g(x_{n-1})}
    \exp\left\{
    -\frac{1}{2} \sum_{n=1}^N \left(
      \frac{\Delta x_n - f(x_{n-1})\Delta t_n}{g(x_{n-1})\sqrt{\Delta t_n}}
      \right)^2
    \right\} \\
  &=:
\int \D{x} F[x] J\exp\left\{
  -\frac{1}{2} \int \mathcal{L}(x(t), \dot{x}(t)) \dd{t}
\right\},
\end{align}
where $J$ is a Jacobian factor which depends on $g$ but not $f$ and the
Lagrangian is given by
\begin{equation}
\mathcal{L}(x, \dot x) = \left( \frac{\dot{x} - f(x)}{g(x)} \right)^2.
\end{equation}
The derivative of $x$ here may be thought of distributionally; however, the
expression for the measure $\D{x} \exp\{-\frac{1}{2}\int \mathcal{L}\}$ need
only be defined on the Cameron--Martin subspace\footnote{The Hilbert space of functions
  whose derivatives are square integrable. The Wiener space is the space of
  continuous functions. This is not a Hilbert space, instead it is a Banach space
  with the supremum norm. The measure of the Cameron--Martin space is in fact
  zero; the sample paths of the SDE are almost-surely nowhere differentiable.};
guaranteeing its unique extension to the whole Wiener space. In contrast to
this, the expression inside the expectation $F[x]$ should be defined
almost-everywhere, and hence care should be taken when $\dot{x}$ appears in an
expectation.

We may expand the quadratic to separate the Lagrangian into a drift-free term
plus terms dependent on the drift
\begin{align}
\mathcal{L} &= \left( \frac{\dot{x}}{g(x)} \right)^2 \label{eq:lagrange}\\
&- 2 f^T(x) g^{-2}(x) \dot{x} \\
&+ f^T(x) g^{-2}(x) f(x).
\end{align}
The first term, on line~\ref{eq:lagrange}, is the Lagrangian for the process
$Y$, which has no drift term. Shuffling the rest of the terms out of the measure
and into the expectation, we have
\begin{equation}
\E_{X}\bigl[ F[X] \bigr] =
\int \D{x} F[x] e^{S[x]} \exp\left\{
  -\frac{1}{2} \int \left( \frac{\dot{x}}{g(x)} \right)^2 \dd{t}
\right\},
\end{equation}
where
\begin{align}
S[X] &= \int f^T(X_t) g^{-2}(X_t) \dot{X}_t \dd{t} \\
     &\quad -\frac{1}{2} \int f^T(X_t) g^{-2}(X_t) f(X_t) \dd{t} \\
     &=\int f^T(X_t) g^{-2}(X_t) \dd{X_t} \\
     &\quad -\frac{1}{2} \int f^T(X_t) g^{-2}(X_t) f(X_t) \dd{t}.
\end{align}
Note that on the third line we have used the distributional definition of
$\dot{X}$; i.e.\ $\dot{X}_t \dd{t} = \dd X_t$.

Lastly, using the definition of conditional expectation, we have that
\begin{equation}
\E_A\Bigl[ F[A] \, p(B=b \vert A) \Bigr] = 
\E_A\Bigl[ p(B=b \vert A) \Bigr]
\E_A\Bigl[ F[A] \,\Big\vert\, B=b \Bigr].
\end{equation}
By trivial substitution, we have
\begin{equation}
p_X(x_{1:N}) = p_Y(x_{1:N}) \E_{Y} \left[
  e^{S[Y]} \,\middle\vert\, \{Y(t_n) = x_n\}_{n=1}^{N}
\right].
\end{equation}

\section{Sampling Bridges}
\label{app:bridge}

Let $Y$ be a Gauss--Markov process with kernel $k$. Since the
process is Markov, we need only sample the process conditioned on two
end-points. Let $y = (y_0,y_T)$ be the known values of the process at times
$\tau=(t_0,T)$. Let $\tilde Y$ be the process defined by
\begin{equation}
\tilde{Y}_t := Y_t + K(t, \tau) K(\tau, \tau)^{-1} \bigl( y - Y_\tau \bigr),
\quad
\mbox{for}
\quad
t_0 \le t \le T,
\end{equation}
where $K$ is the matrix whose entries are $k(t_i, t_j)$.
This distribution of $\tilde Y$ is the same as the distribution of $Y$
conditioned on the values $y$ at times $\tau$. Note that once
$Y_t$ has been sampled --- which may require a cumulative sum --- this
transformation can be performed independently, and hence in parallel, for each
$t$. For each consecutive pair of observations, the bridge process can be
sampled independently.

In the simplest case, where $Y$ is a Brownian motion starting from $y_0$ at time
$t_0 = 0$, this reduces to
\begin{align}
  \tilde{Y}_t
  &= Y_t + \tfrac{k(t,T)}{k(T,T)} (y_T - Y_T) \\
  &= Y_t + \tfrac{t}{T} (y_T - Y_T).
\end{align}
And similarly for any other processes which can be sampled with a fixed starting
point.

In the more general case we can invert the two by two matrix to get an explicit
formula for $\tilde{Y}_t$
\begin{align}
  \tilde{Y}_t
  = Y_t &+ \frac{k(t,t_0)k(T,T) - k(t,T)k(t_0,T)}{k(t_0,t_0)k(T,T) - k(t_0,T)^2} (y_0 - Y_0) \\
   &+\frac{k(t,T)k(t_0,t_0) - k(t,t_0)k(t_0,T)}{k(t_0,t_0)k(T,T) - k(t_0,T)^2} (y_T - Y_T).
\end{align}
This may be extended straightforwardly to other cases, such as conditioning on
the values of the integral, or derivative of the process etc.

\section{Experiments}
\label{app:exp}

\subsection{Probability estimates}

\begin{figure}[htb]
  \centering
  \includegraphics[width=0.5\textwidth]{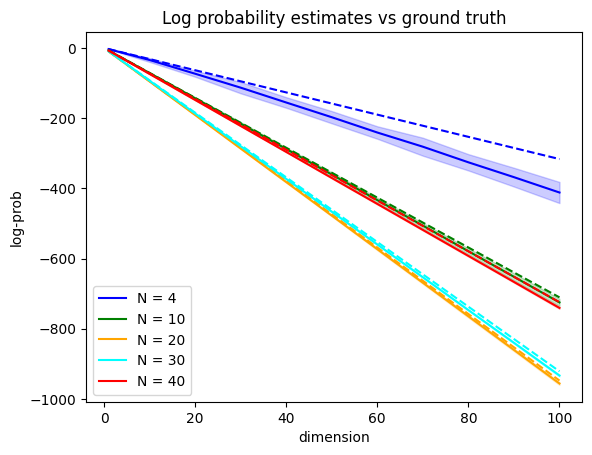}
  \caption{\label{fig:ou} Log-probability estimates for an Ornstein--Uhlenbeck
    process. The solid lines indicate the mean value of the estimator, and the
    shaded regions indicate three standard deviations from the mean. The dashed
    lines are the analytic log-probability values.}
\end{figure}

As a sanity check, we compare the log-probability estimates of our algorithm to
the actual log-probabilities of an Ornstein--Uhlenbeck process. The probability
density for this process is available in closed form for any number of
observations.
We estimate the probability of a set of data points evenly spaced over a time
interval of $[0, 10]$, taken from a sine function
$x_n=\sin(\nicefrac{\pi}{10}\times t_n)$.
For these estimates we used $1000$ time steps to estimate the integral $S$, and
averaged over $100$ trajectories, independently sampled from a Brownian bridge
passing through the observations.
We repeated this process $1000$ times and report the mean and standard deviation
of these estimates.
Figure~\ref{fig:ou} shows these estimates compared to the analytic	
probability given by the Ornstein--Uhlenbeck process for various numbers of
observations as a function of the dimension of the problem.

An interesting observation is that the variance of the estimator seems to
decrease with the number of data points. This is atypical of probability
estimators, since the magnitude of the log-probability increases linearly with
the number of observations (not to be confused with the number of samples used
to estimate the log-probability).
In almost all cases, with the exception of 4 observations in more than $40$
dimensions, our algorithm gives accurate log-probability estimates with small
standard deviations. The error of this exception is approximately 30\%, but is
significantly reduced by adding more observations, or using more samples. We
believe this error is acceptable considering the small number of observations
and the large scale.
For 10 observations in $100$ dimensions the standard deviation of our
estimator is only 0.4\% of the magnitude of the log-probability and the error of
the mean estimate is 2\% of the analytic log-probability. This small bias can be
further reduced by using more samples or a smaller $\Delta t$.
A single estimate in the $100$ dimensional case took less than half a second to
compute on a laptop CPU, and is therefore feasible to use in the inner loop of a
learning algorithm.

\subsection{Lorenz}
\subsubsection{Data set}

We generated data by simulating a Lorenz system with the
exact same parameters as \citet{NeuralSDE}. We generated 16 independent
trajectories, with initial values sampled from a standard normal distribution,
and discretely sampled the values of the process over 200 equally spaced time
points ranging from 0 to 10. Unlike \citet{NeuralSDE}, we did not add any
observation noise to the data.

We deliberately used a sparser data set than \citet{NeuralSDE}, due to personal
experience of the difficulty of fitting SDE models on sparse observations.

\subsubsection{Model}

The model we used to fit to the observations was
\begin{equation}
\dd{X}_t = f_{\theta}(X_t) \dd{t} + \sigma \,\dd W,
\end{equation}
where $\sigma$ is a scalar and $f_{\theta}$ is a fully-connected neural network
with ReLU nonlinearities and parameters $\theta$. We used layer sizes of $3 \to
32 \to 256 \to 32 \to 3$.

\subsubsection{Training parameters}

We used precisely the same parameters for each model, the only difference in the
training procedures being the loss functions used. In the first case, the loss
function per trajectory was
\begin{equation}
  L_{\mathrm{mse}}(\theta ; x_{1:N}) = \frac{1}{K N} \sum_{k,n} \bigl(
  X_{t_n}^{(k)} - x_{n}
  \bigr)^2,
\end{equation}
where $N=200$ is the number of observations per trajectory, and $K$ is the
number of sampled paths used to estimate the MSE, and $X^{(k)}$ is sampled using
an SDE integrator starting at $X_{t_1}^{(k)} = x_1$. We then averaged this loss
function over the 16 independent observed trajectories.

In the second case, the loss function per trajectory was
\begin{equation}
L_{\mathrm{PI}}(\theta ; x_{1:N}) = -\frac{1}{N} \log\hat{p}_{X}(x_{1:N} \,\vert\, \theta),
\end{equation}
where $\hat{p}_X(x_{1:N} \,\vert\, \theta)$ is calculated using
Algorithm~\ref{alg:diff-ind} with $K$ trajectory samples. Again this loss was
averaged over the 16 independent observed trajectories. In both cases we used
$K=64$ and a time step size of $\Delta t$ = $10^{-2}$.

For each model we used the Adam optimizer with a learning rate of $10^{-3}$ and
ran the optimization algorithm for $10^4$ iterations.

\subsection{Variational Gaussian processes}

Our approach allows more straightforward implementation of SDE models using a
Gaussian process~(GP) prior for the drift function. Some previous approaches
often resort to approximations such as using only the predictive mean or MAP
estimates~\citep{LearningStochasticDifferential}. For dense observations,
directly discretizing the SDE and fitting the GP to the observation deltas may
be viable. For sparse observations this is much more difficult. Our proposed
method makes variational GP methods slightly simpler, since the values of
$f(Y_t)$ can be sampled jointly along the path. However, it is important to note
that various complexities still arise such as the difficulty of efficiently
sampling high-dimensional Gaussian distributions.

We used the kernel interpolation method of \citet{KissGP} which stores a grid of
inducing locations per dimension. Storing inducing locations per dimension for a
Kronecker product kernel effectively gives a variational distribution over
$m^d$ inducing points with only $d \times m$ memory requirements, where $m$ is
the number of inducing points per dimension and $d$ is the dimension. When these
inducing locations are evenly spaces, the covariance factors have a Toeplitz
structure which allows sampling in linearithmic time complexity.

To fit the model we used gradient ascent on the
ELBO which is calculated as follows:
\begin{equation}
  \textsc{Elbo} = \E_{f\sim q}\bigl[\log{p}(x|f)\bigr] - \operatorname{KL}\bigl[q(f)|p(f)\bigr].
\end{equation}
The KL divergence is between two Gaussian distributions calculated at the
inducing locations in the usual way. For the likelihood term, we generated
trajectories (as in Algorithm~\ref{alg:diff-ind}), followed by jointly sampling
the variational GP posterior along these trajectories and proceed with the rest
of the computation in Algorithm~1 using these samples. Our optimization
objective is calculated
\begin{equation}
  L = - \frac{1}{K}\sum_K \log\hat{p}(x_{1:N}\,|\,f^{(k)}) + \operatorname{KL}\bigl[q(f)|p(f)\bigr],
\end{equation}
where each $f^{(k)}$ is an independent sample over the whole path of $Y$. In
other words, $f^{(k)}(Y_{t_1})$ and $f^{(k)}(Y_{t_2})$ are correlated, but
$f^{(k)}(Y_{t_1})$ and $f^{(j)}(Y_{t_2})$ are not when $k \ne j$.

\end{document}